\documentclass[11pt]{article}
\usepackage{coling2016}
\usepackage{times}
\usepackage{url}
\usepackage{latexsym}
\usepackage{graphicx}
\usepackage{tablefootnote}

\title{Categorization of Semantic Roles for Dictionary Definitions}

\author{Vivian S. Silva, Siegfried Handschuh and Andr\'{e} Freitas\\
 Department of Computer Science and Mathematics \\ University of Passau \\ 
 Innstra\ss{}e 43, 94032, Passau, Germany \\
 {\tt {\small \{vivian.santossilva, siegfried.handschuh, andre.freitas\}@uni-passau.de}}}

\date{}

\begin{document}
\maketitle
\begin{abstract}
Understanding the semantic relationships between terms is a fundamental task in natural language processing applications. While structured resources that can express those relationships in a formal way, such as ontologies, are still scarce, a large number of linguistic resources gathering dictionary definitions is becoming available, but understanding the semantic structure of natural language definitions is fundamental to make them useful in semantic interpretation tasks. Based on an analysis of a subset of WordNet's glosses, we propose a set of semantic roles that compose the semantic structure of a dictionary definition, and show how they are related to the definition's syntactic configuration, identifying patterns that can be used in the development of information extraction frameworks and semantic models.
\end{abstract}

\section{Introduction}\label{sec:intro}\blfootnote{This work is licensed under a Creative Commons Attribution 4.0 International Licence. Licence	details: http://creativecommons.org/licenses/by/4.0/}
Many natural language understanding tasks such as Text Entailment and Question Answering systems are dependent on the interpretation of the semantic relationships between terms. The challenge on the construction of robust semantic interpretation models is to provide a model which is both comprehensive (capture a large set of semantic relations) and fine-grained. While semantic relations (high-level binary predicates which express relationships between words) can serve as a semantic interpretation model, in many cases, the relationship between words cannot be fully articulated as a single semantic relation, depending on a contextualization that involves one or more target words, their corresponding semantic relationships and associated logical operators (e.g. modality, functional operators).

Natural language definitions of terms, such as dictionary definitions, are resources that are still underutilized in the context of semantic interpretation tasks. The high availability of natural language definitions in different domains of discourse, in contrast to the scarcity of comprehensive structured resources such as ontologies, make them a candidate linguistic resource to provide a data source for fine-grained semantic models. 

Under this context, understanding the syntactic and semantic ``shape'' of natural language definitions, i.e., how definitions are usually expressed, is fundamental for the extraction of structured representations and for the construction of semantic models from these data sources. This paper aims at filling this gap by providing a systematic analysis of the syntactic and semantic structure of natural language definitions and proposing a set of semantic roles for them. By \textit{semantic role} here we mean entity-centered roles, that is, roles representing the part played by an expression in a definition, showing how it relates to the entity being defined. WordNet \cite{fellbaum1998wordnet}, one of the most employed linguistic resources in semantic applications,  was used as a corpus for this task. The analysis points out the syntactic and semantic regularity of definitions, making explicit an enumerable set of syntactic and semantic patterns which can be used to derive information extraction frameworks and semantic models.

The contributions of this paper are: (i) a systematic preliminary study of syntactic and semantic relationships expressed in a corpus of definitions, (ii) the derivation of semantic categories for the classification of semantic patterns within definitions, and (iii) the description of the main syntactic and semantic shapes present in definitions, along with the quantification of the distribution of these patterns.

The paper is organized as follows: Section \ref{sec:strucasp} presents the basic structural aspects of definitions according to the classic theory of definitions. Section \ref{sec:semrole} introduces the proposed set of semantic roles for definitions. Section \ref{sec:ident} outlines the relationship between semantic and syntactic patterns. Section \ref{sec:relwork} lists related work, followed by the conclusions and future work in Section \ref{sec:concl}.

\section{Structural Aspects of Definitions}\label{sec:strucasp}
Swartz \shortcite{swartz2007definitions} describe lexical, or dictionary definitions as reports of common usage (or usages) of a term, and argue that they allow the improvement and refinement of the use of language, because they can be used to increase vocabulary (introducing people to the meaning and use of words new to them), to eliminate certain kinds of ambiguity and to reduce vagueness. A clear and properly structured definition can also provide the necessary identity criteria to correctly allocate an entity in an ontologically well-defined taxonomy \cite{guarino2002evaluating}.

Some linguistic resources, such as WordNet, organize concepts in a taxonomy, so the genus-differentia definition pattern would be a suitable way to represent the subsumption relationship among them. The genus and differentia concepts date back to Aristotle's writings concerning the theory of definition \cite{berg1982aristotle,granger1984aristotle,lloyd1962genus} and are most commonly used to describe entities in the biology domain, but they are general enough to define concepts in any field of knowledge. An example of a genus-differentia based definition is the Aristotelian definition of a human: ``a human is a rational animal''. \textit{Animal} is the genus, and \textit{rational} is the differentia distinguishing humans from other animals.

Another important aspect of the theory of definition is the distinction between essential and non-essential properties. As pointed by Burek \shortcite{burek2004adoption}, stating that ``a human is an animal'' informs an essential property for a human (being an animal), but the sentence ``human is civilized'' does not communicate a fundamental property, but rather something that happens to be true for humans, that is, an incidental property.

Analyzing a subset of the WordNet definitions to investigate their structure, we noticed that most of them loosely adhere to the classical theory of definition: with the exception of some samples of what could be called ill-formed definitions, in general they are composed by a linguistic structure that resembles the genus-differentia pattern, plus optional and variable incidental properties. Based on this analysis, we derived a set of semantic roles representing the components of a lexical definition, which are described next.

\section{Semantic Roles for Lexical Definitions}\label{sec:semrole}
Definitions in WordNet don't follow a strict pattern: they can be constructed in terms of the entity's immediate superclass or rather using a more abstract ancestral class. For this reason, we opted for using the more general term \textbf{supertype} instead of the classical \textit{genus}. A supertype is either the immediate entity's superclass, as in ``footwear: \textit{clothing} worn on a person's feet'', being \textit{footwear} immediately under \textit{clothing} in the taxonomy; or an ancestral, as in ``illiterate: a \textit{person} unable to read'', where \textit{illiterate} is three levels below \textit{person} in the hierarchy.

Two different types of distinguishing features stood out in the analyzed definitions, so the differentia component was split into two roles: \textbf{differentia quality} and \textbf{differentia event}. A differentia quality is an essential, inherent property that distinguishes the entity from the others under the same supertype, as in ``baseball\_coach: a coach \textit{of baseball players}''. A differentia event is an action, state or process in which the entity participates and that is mandatory to distinguish it from the others under the same supertype. It is also essential and is more common for (but not restricted to) entities denoting roles, as in ``roadhog: a driver \textit{who obstructs others}''.

As any expression describing events, a differentia event can have several subcomponents, denoting time, location, mode, etc. Although many roles could be derived, we opted to specify only the ones that were more recurrent and seemed to be more relevant for the definitions' classification: \textbf{event time} and \textbf{event location}. Event time is the time in which a differentia event happens, as in ``master\_of\_ceremonies: a person who acts as host \textit{at formal occasions}''; and event location is the location of a differentia event, as in ``frontiersman: a man who lives \textit{on the frontier}''.

A \textbf{quality modifier} can also be considered a subcomponent of a differentia quality: it is a degree, frequency or manner modifier that constrain a differentia quality, as in ``dart: run or move \textit{very} quickly or hastily'', where \textit{very} narrows down the differentia quality \textit{quickly} associated to the supertypes \textit{run} and \textit{move}.

The \textbf{origin location} role can be seen as a particular type of differentia quality that determines the entity's location of origin, but in most of the cases it doesn't seem to be an essential property, that is, the entity only happens to occur or come from a given location, and this fact doesn't account to its essence, as in ``Bartramian\_sandpiper: large plover-like sandpiper \textit{of North American fields and uplands}'', where \textit{large} and \textit{plover-like} are essential properties to distinguish \textit{Bartramian\_sandpiper} from other sandpipers, but occurring in \textit{North American fields and uplands} is only an incidental property.

The \textbf{purpose} role determines the main goal of the entity's existence or occurrence, as in ``redundancy: repetition of messages \textit{to reduce the probability of errors in transmission}''. A purpose is different from a differentia event in the sense that it is not essential: in the mentioned example, a repetition of messages that fails to reduce the probability of errors in transmission is still a redundancy, but in ``water\_faucet: a faucet \textit{for drawing water} from a pipe or cask”'', \textit{for drawing water is} a differentia event, because a faucet that fails this condition is not a water faucet.

Another event that is also non-essential, but rather brings only additional information to the definition is the \textbf{associated fact}, a fact whose occurrence is/was linked to the entity's existence or occurrence, accounting as an incidental attribute, as in ``Mohorovicic: Yugoslav geophysicist \textit{for whom the Mohorovicic discontinuity was named}''.

Other minor, non-essential roles identified in our analysis are: \textbf{accessory determiner}, a determiner expression that doesn't constrain the supertype-differentia scope, as in ``camas: \textit{any of several} plants of the genus Camassia'', where the expression \textit{any of several} could be removed without any loss in the definition meaning; \textbf{accessory quality}, a quality that is not essential to characterize the entity, as in ``Allium: \textit{large} genus of perennial and biennial pungent bulbous plants'', where \textit{large} is only an incidental property; and \textbf{[role] particle}, a particle, such as a phrasal verb complement, non-contiguous to the other role components, as in ``unstaple: take the staples \textit{off}'', where the verb \textit{take off} is split in the definition, being \textit{take} the supertype and \textit{off} a supertype particle.

\begin{figure*}[h]
	\centering
	\includegraphics[width=6.3in, height=3.18in]{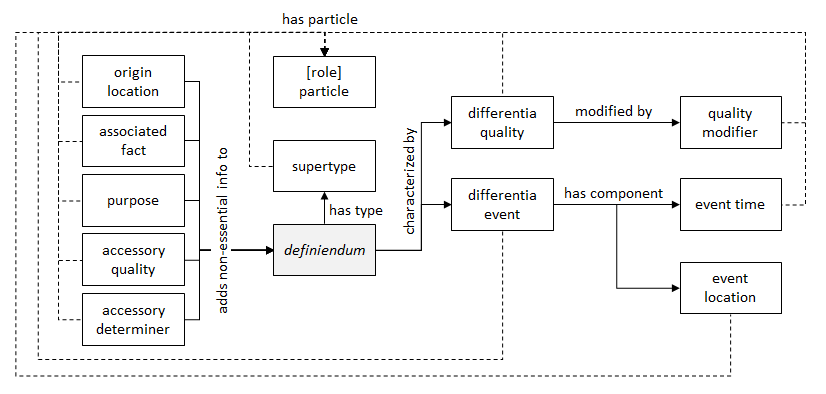}
	\caption{Conceptual model for the semantic roles for lexical definitions. Relationships between \textit{[role] particle} and every other role in the model are expressed as dashed lines for readability.}
	\label{fig:sem_model}
\end{figure*}

The conceptual model in Figure \ref{fig:sem_model} shows the relationship among roles, and between roles and the \textit{definiendum}, that is, the entity being defined.

\section{Identifying Semantic Roles in Definitions}\label{sec:ident}
Once the relevant semantic roles were identified in the manual analysis, the following question emerged: is it possible to extend this classification to the whole definitions database through automated Semantic Role Labelling? Although most SRL systems rely on efficient machine learning techniques \cite{palmer2010semantic}, an initial, preferably large, amount of annotated data is necessary for the training phase.

Since manual annotation is expensive, an alternative would be a rule-based mechanism to automatically label the definitions, based on their syntactic structure, followed by a manual curation of the generated data. As shown in an experimental study by Punyakanok et al. \shortcite{punyakanok2005necessity}, syntactic parsing provides fundamental information for event-centered SRL, and, in fact, this is also true for entity-centered SRL.

\begin{figure*}[h]
	\centering
	\includegraphics[width=6in, height=0.6in]{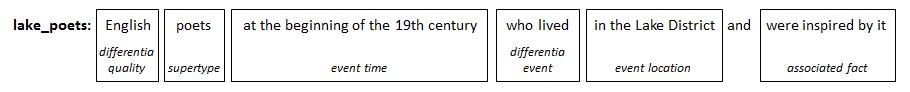}
	\caption{Example of role labeling for the definition of the ``lake\_poets'' synset.}
	\label{fig:classif_example}
\end{figure*}

To draw the relationship between syntactic and semantic structure (as well as defining the set of relevant roles described earlier), we randomly selected a sample of 100 glosses from the WordNet nouns+verbs database\footnote{Adjectives and adverbs are not organized in a taxonomy in WordNet, so are less likely to follow a supertype-differentia pattern, probably requiring a different classification strategy}, being 84 nouns and 16 verbs (the verb database size is only approximately 17\% of the noun database size). First, we manually annotated each of the glosses, assigning to each segment in the sentence the most suitable role. Example sentences and parentheses were not included in the classification. Figure \ref{fig:classif_example} shows an example of annotated gloss. Then, using the Stanford parser \cite{manning2014stanford}, we generated the syntactic parse trees for all the 100 glosses and compared the semantic patterns with their syntactic counterparts.

Table \ref{tab:sem_patterns} shows the distribution of the semantic patterns for the analyzed sample. As can be seen, \textit{(supertype) (differentia quality)} and \textit{(supertype) (differentia event)} are the most frequent patterns, but many others are composed by a combination of three or more roles, usually the supertype, one or more differentia qualities and/or differentia events, and any of the other roles. Since most of them occurred only once (29 out of 42 identified patterns), it is easier to analyze the roles as independent components, regardless of the pattern where they appear. The context can always give some hint about what a role is, but we would expect the role's main characteristics not to change when their ``companions'' in the sentence varies. The conclusions are as follows\footnote{POS tags and non-terminal symbols lists can be found at https://goo.gl/8ndYCw and  https://goo.gl/kuJEc2, respectively}, and are summarized in Table \ref{tab:synt_patterns}:

\begin{table}[h]
	\centering
	\begin{tabular}{|l|r|}
		\hline
		\textbf{Pattern}        & \multicolumn{1}{l|}{\textbf{Total}}  \\ \hline
		(supertype) (differentia quality)                        & 27  \\ \hline
		(supertype) (differentia event)                          & 13  \\ \hline
		(differentia quality) (supertype)                        & 6   \\ \hline
		(supertype) (differentia event) (event location)         & 5   \\ \hline
		(supertype) (differentia quality) (purpose)              & 3   \\ \hline
		(accessory determiner) (supertype) (differentia event)   & 3   \\ \hline
		(accessory determiner) (supertype) (differentia quality) & 2   \\ \hline
		(supertype) OR(differentia quality)+                     & 2   \\ \hline
		(supertype) (origin location)                            & 2   \\ \hline
		(differentia quality) (supertype) (differentia quality)  & 2   \\ \hline
		OR(supertype)+ (differentia event)                       & 2   \\ \hline
		(differentia quality)+ (supertype)                       & 2   \\ \hline
		(differentia quality)+ (supertype) (differentia event)   & 2   \\ \hline
		Other                                                    & 29  \\ \hline
		\textbf{Total}                                 & \textbf{100}  \\ \hline
	\end{tabular}
	\caption{Distribution of semantic patterns for the analyzed definitions. ``Other'' refers to patterns that ocurred only once. (\textit{role})+ indicated the occurrence of two or more consecutive instances of the role, and OR(\textit{role})+ indicates the same, but with the conjunction ``or'' connecting the instances.}
	\label{tab:sem_patterns}
\end{table}

\textbf{Supertype:} it's mandatory in a well-formed definition, and indeed 99 out of the 100 sentences analyzed have a supertype (the gloss for \textit{Tertiary\_period} -- ``from 63 million to 2 million years ago'' – lacks a supertype and could, then, be considered an ill-formed definition). For verbs, it is the leftmost VB and, in some cases, subsequent VBs preceded by a CC (``or'' or ``and''). This is the case whenever the parser correctly classifies the gloss' head word as a verb (11 out of 16 sentences). For nouns, in most cases (70 out of 83) the supertype is contained in the innermost and leftmost NP containing at least one NN. It is the whole NP (discarding leading DTs) if it exists as an entry in WN, or the largest rightmost sequence that exists in WN otherwise. In the last case, the remaining leftmost words correspond to one or more differentia qualities. If the NP contains CCs, more than one supertype exist, and can be identified following the same rules just described. The 13 sentences that don't fit this scenario include some non-frequent grammatical variations, parser errors and the presence of accessory determiners, described later.

\textbf{Differentia quality:} for verbs, this is the most common identifying component in the definition. It occurs in 14 out of the 16 sentences. The other two ones are composed by a single supertype (that would better be seen as a synonym), and by a conjunction of two supertypes. The differentia quality is usually a PP (5 occurrences) or a NP (4 occurrences) coming immediately after the supertype. JJs inside ADJPs (3 occurrences) or RBs inside ADVPs (1 occurrence) are also possible patterns, where the presence of CCs indicates the existence of more than one differentia quality. For nouns, two scenarios outstand: the differentia quality preceding the supertype, where it is composed by the leftmost words in the same NP that contains the supertype but are not part of the supertype itself, as described above; and the differentia quality coming after the supertype, predominantly composed by a PP, where the prevailing introductory preposition is ``of''. These two scenarios cover approximately 90\% of all analyzed sentences where one or more differentia qualities occur.

\textbf{Differentia event:} differentia events occurs only for nouns, since verbs can't represent entities that can participate in an event (i.e., \textit{endurants} in the ontological view). They are predominantly composed by either an SBAR or a VP (under a simple clause or not) coming after the supertype. This is the case in approximately 92\% of the analyzed sentences where differentia events occur. In the remaining samples, the differentia event is also composed by a VP, but under a PP and immediately after the introductory preposition.

\textbf{Event location:} event locations only occur in conjunction with a differentia event, so they will usually be composed by a PP appearing inside a SBAR or a VP. Being attached to a differentia event helps to distinguish an event location from other roles also usually composed by a PP, but additional characteristics can also provide some clues, like, for example, the presence of named entities denoting locations, such as ``Morocco'' and ``Lake District'', which appear in some of the analyzed glosses.

\textbf{Event time:} the event time role has the same characteristics of event locations: only occurs in conjunction with a differentia event and is usually composed by a PP inside a SBAR or a VP. Again, additional information such as named entities denoting time intervals, for example, ``the 19th century'' in one of the analyzed glosses, is necessary to tell it apart from other roles.

\textbf{Origin location:} origin locations are similar to event locations, but occurring in the absence of an event, so it is usually a PP that does not appear inside a SBAR or a VP and that frequently contains named entities denoting locations, like ``United States'', ``Balkan Peninsula'' and ``France'' in our sample glosses. A special case is the definition of entities denoting instances, where the origin location usually comes before the supertype and is composed by a NP (also frequently containing some named entity), like the definitions for \textit{Charlotte\_Anna\_Perkins\_Gilman} -- ``United States feminist'' -- and \textit{Joseph\_Hooker} -- ``United States general [\dots]'', for example.

\textbf{Quality modifier:} quality modifiers only occur in conjunction with a differentia quality. Though this role wasn't very frequent in our analysis, it is easily identifiable, as long as the differentia quality component has already been detected. A syntactic dependency parsing can show whether some modifier (usually an adjective or adverb) references, instead of the supertype, some of the differentia quality's elements, modifying it.

\textbf{Purpose:} the purpose component is usually composed by a VP beginning with a TO (``to'') or a PP beginning with the preposition ``for'' and having a VP right after it. In a syntactic parse tree, a purpose can easily be mistaken by a differentia event, since the difference between them is semantic (the differentia event is essential to define the entity, and the purpose only provide additional, non-essential information). Since it provides complementary information, it should always occur in conjunction with an identifying role, that is, a differentia quality and/or event. Previously detecting these identifying roles in the definition, although not sufficient, is necessary to correctly assign the purpose role to a definition's segment.

\textbf{Associated fact:} an associated fact has characteristics similar to those of a purpose. It is usually composed by a SBAR or by a PP not beginning with ``for'' with a VP immediately after it (that is, not having the characteristics of a purpose PP). Again, the difference between an associated fact and a differentia event is semantic, and the same conditions and principles for identifying a purpose component also apply.

\textbf{Accessory determiner:} accessory determiners come before the supertype and are easily recognizable when they don't contain any noun, like ``any of several'', for example: it will usually be the whole expression before the supertype, which, in this case, is contained in the innermost and leftmost NP having at least one NN. If it contains a noun, like ``a type of'', ``a form of'', ``any of a class of'', etc., the recognition becomes more difficult, and it can be mistaken by the supertype, since it will be the leftmost NP in the sentence. In this case, a more extensive analysis in the WN database to collect the most common expressions used as accessory determiners is necessary in order to provide further information for the correct role assignment.

\textbf{Accessory quality:} the difference between accessory qualities and differentia qualities is purely semantic. It is usually a single adjective, but the syntactic structure can't help beyond that in the accessory quality identification. Again, the presence of an identifying element in the definition (preferably a differentia quality) associated with knowledge about most common words used as accessory qualities can provide important evidences for the correct role detection.

\textbf{[Role] particle:} although we believe that particles can occur for any role, in our analysis it was very infrequent, appearing only twice and only for supertypes. It is easily detectable for phrasal verbs, for example, \textit{take off} in ``take the staples off'', since the particle tends to be classified as PRT in the syntactic tree. For other cases, it is necessary a larger number of samples such that some pattern can be identified and a suitable extraction rule can be defined.

\begin{table}[h]
	\centering
	\begin{tabular}{|l|l|}
		\hline
		\textbf{Role}        & \textbf{Most common syntactic patterns}   \\ \hline
		Supertype            & innermost and leftmost NP containing at least one NN            \\ \hline
		Differentia quality  & leftovers\tablefootnote{Words that are not part of the largest sequence in the NP found as an entry in WN} in the innermost and leftmost NP; PP beginning with ``of''' \\ \hline
		Differentia event    & SBAR; VP		  \\ \hline
		Event location       & PP inside a SBAR or VP, possibly having a location named entity        \\ \hline
		Event time           & PP inside a SBAR or VP, possibly having a time interval named entity  \\ \hline
		Origin location      & PP not inside a SBAR or VP, possibly having a location named entity  \\ \hline
		Quality modifier     & NN, JJ or RB referring to an element inside a differentia quality       \\ \hline
		Purpose              & VP beginning with TO; PP beginning with ``for'' with a VP right after   \\ \hline
		Associated fact      & SBAR; PP not beginning with ``for'' with a VP right after         \\ \hline
		Accessory determiner & whole expression before supertype; common accessory expression    \\ \hline
		Accessory quality    & JJ, presence of a differentia quality, common accessory word          \\ \hline
		{[}Role{]} particle  & PRT          \\ \hline
	\end{tabular}
	\caption{Most common syntactic patterns for each semantic role.}
	\label{tab:synt_patterns}
\end{table}

\section{Related Work}\label{sec:relwork}
The task described in this work is a form of Semantic Role Labeling (SRL), but centered on entities instead of events. Typically, SRL has as primary goal to identify what semantic relations hold among a predicate (the main verb in a clause) and its associated participants and properties \cite{marquez2008semantic}. Focusing on determining ``who'' did ``what'' to ``whom'', ``where'', ``when'', and ``how'', the labels defined for this task include \textit{agent}, \textit{theme}, \textit{force}, \textit{result} and \textit{instrument}, among others \cite{jurafsky2000speech}. 

Liu and Ng \shortcite{liu2007learning} perform SRL focusing on nouns instead of verbs, but most noun predicates in NomBank, which were used in the task, are verb nominalizations. This leads to the same event-centered role labeling, and the same principles and labels used for verbs apply.

Kordjamshidi et al. \shortcite{kordjamshidi2010spatial} describe a non-event-centered semantic role labeling task. They focus on spatial relations between objects, defining roles such as \textit{trajectory}, \textit{landmark}, \textit{region}, \textit{path}, \textit{motion}, \textit{direction} and \textit{frame of reference}, and develop an approach to annotate sentences containing spatial descriptions, extracting topological, directional and distance relations from their content.

Regarding the structural aspects of lexical definitions, Bodenreider and Burgun \shortcite{bodenreider2002characterizing} present an analysis of the structure of biological concept definitions from different sources. They restricted the analysis to anatomical concepts to check to what extent they fit the genus-differentia pattern, the most common method used to classify living organisms, and what the other common structures employed are, in the cases where that pattern doesn't apply.

Burek \shortcite{burek2004adoption} also sticks to the Aristotelian classic theory of definition, but instead of analyzing existing, natural language definitions, he investigates a set of ontology modeling languages to examine their ability to adopt the genus-differentia pattern and other fundamental principles, such as the essential and non-essential property differentiation, when defining a new ontology concept by means of axioms, that is, in a structured way rather than in natural language. He concludes that Description Logic (DL), Unified Modeling Language (UML) and Object Role Modeling (ORM) present limitations to deal with some issues, and proposes a set of \textit{definitional tags} to address those points.

The information extraction from definitions has also been widely explored with the aim of constructing structured knowledge bases from machine readable dictionaries \cite{vossen1992automatic,calzolari1991acquiring,copestake1991lkb,vossen1991converting,vossen1994untangling}. The development of a Lexical Knowledge Base (LKB) also used to take into account both semantic and syntactic information from lexical definitions, which were processed to extract the definiendum's genus and differentiae. To populate the LKB, typed-feature structures were used to store the information from the differentiae, which were, in turn, transmitted by inheritance based on the genus information. A feature structure can be seen as a set of attributes for a given concept, such as ``origin'', ``color'', ``smell'', ``taste'' and ``temperature'' for the concept \textit{drink} (or for a more general concept, such as \textit{substance}, from which \textit{drink} would inherit its features), for example, and the differentiae in a definition for a particular drink would be the values that those features assume for that drink, for example, ``red'', ``white'', ``sweet'', ``warm'', etc. As a result, concepts could be queried using the values of features as filters, and words defined in different languages could be related, since they were represented in the same structure. To build the feature structures, restricted domains covering subsets of the vocabulary were considered, since having every relevant attribute for every possible entity defined beforehand is not feasible, being more overall strategies required in order to process definitions in large scale.

\section{Conclusion}\label{sec:concl}
We proposed a set of semantic roles that reflect the most common structures of dictionary definitions. Based on an analysis of a random sample composed by 100 WordNet noun and verb glosses, we identified and named the main semantic roles and their compositions present on dictionary definitions. Moreover, we compared the identified semantic patterns with the definitions' syntactic structure, pointing out the features that can serve as input for automatic role labeling. The proposed semantic roles list is by no means definitive or exhaustive, but a first step at highlighting and formalizing the most relevant aspects of widely used intensional level definitions.

As future work, we intend to implement a rule-based classifier, using the identified syntactic patterns to generate an initial annotated dataset, which can be manually curated and subsequently feed a machine learning model able to annotate definitions in large scale. We expect that, through a systematic classification of their elements, lexical definitions can bring even more valuable information to semantic tasks that require world knowledge.

\section*{Acknowledgements}
Vivian S. Silva is a CNPq Fellow -- Brazil.

\bibliography{cogalex2016}
\bibliographystyle{acl}

\end{document}